\documentclass{article}

\usepackage{microtype}
\usepackage{graphicx}
\usepackage{subfigure}
\usepackage{booktabs} 

\usepackage{hyperref}

\usepackage{amsmath}
\usepackage{amssymb}
\usepackage{dsfont}
\usepackage{amsthm}
\usepackage{mathtools}
\usepackage{stfloats}
\usepackage{multirow}

\newtheorem{theorem}{Theorem}

\newcommand{\eg}{\emph{e.g. }}

\newcommand{\ie}{\emph{i.e. }}
\newcommand{\etc}{\emph{etc}}
\newcommand{\wrt}{\emph{w.r.t. }}

\usepackage[accepted]{icml2020}

\icmltitlerunning{Training Binary Neural Networks through Learning with Noisy Supervision}

\begin{document}

\twocolumn[
\icmltitle{Training Binary Neural Networks through Learning with Noisy Supervision}




\begin{icmlauthorlist}
\icmlauthor{Kai Han}{keylab,huawei}
\icmlauthor{Yunhe Wang}{huawei}
\icmlauthor{Yixing Xu}{huawei}
\icmlauthor{Chunjing Xu}{huawei}
\icmlauthor{Enhua Wu}{keylab,um}
\icmlauthor{Chang Xu}{sydney}
\end{icmlauthorlist}

\icmlaffiliation{keylab}{State Key Lab of Computer Science, Institute of Software, CAS \& University of Chinese Academy of Sciences}
\icmlaffiliation{huawei}{Noah's Ark Lab, Huawei Technologies}
\icmlaffiliation{um}{University of Macau}
\icmlaffiliation{sydney}{School of Computer Science, Faculty of Engineering, University of Sydney}

\icmlcorrespondingauthor{Yunhe Wang}{yunhe.wang@huawei.com}
\icmlcorrespondingauthor{Enhua Wu}{ehwu@umac.mo}
\icmlcorrespondingauthor{Chang Xu}{c.xu@sydney.edu.au}

\icmlkeywords{Binary Neural Networks, Convolutional Neural Networks, Learning with Noisy Labels}

\vskip 0.3in
]



\printAffiliationsAndNotice{}  

\begin{abstract}
	This paper formalizes the binarization operations over neural networks from a learning perspective. In contrast to classical hand crafted rules (\eg hard thresholding) to binarize full-precision neurons, we propose to learn a mapping from full-precision neurons to the target binary ones. Each individual weight entry will not be binarized independently. Instead, they are taken as a whole to accomplish the binarization, just as they work together in generating convolution features. To help the training of the binarization mapping, the full-precision neurons after taking sign operations is regarded as some auxiliary supervision signal, which is noisy but still has valuable guidance.  An unbiased estimator is therefore introduced to mitigate the influence of the supervision noise. Experimental results on benchmark datasets indicate that the proposed binarization technique attains consistent improvements over baselines.
\end{abstract}

\section{Introduction}

\begin{figure}[!thb]
	\centering
	\includegraphics[width=0.95\linewidth]{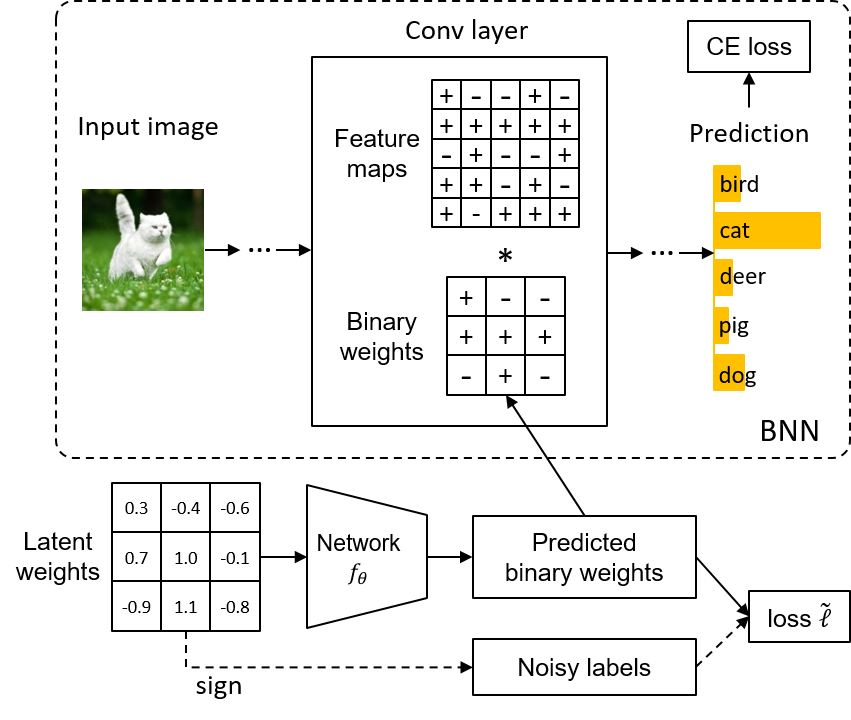}
	\vspace{-1.0em}
	\caption{Framework of learning binary neurons with noisy supervision. A network $f$ is utilized to predict the binary weights with supervision of the noisy labels obtained by sign function.}
	\label{fig:framework}
	\vspace{-1.5em}
\end{figure}

Deep convolutional neural networks (CNNs) have achieved much success in many real-world applications such as image recognition~\cite{resnet,a3m}, object detection~\cite{fasterrcnn}, and semantic segmentation~\cite{deeplab}. These CNN models usually consume high computational resource, and thus they cannot be easily deployed on embedded devices. A series of model compression and acceleration methods~\cite{deepcompression,addernet} have been proposed to reduce the number of parameters and FLOPs of CNNs, including network pruning~\cite{deepcompression,l1-pruning,shu2019co}, tensor decomposition~\cite{denton2014exploiting}, knowledge distillation~\cite{Distill,datafree}, efficient model design~\cite{mobilenet,ghostnet}, and model quantization~\cite{gupta2015deep,bnn}. These methods have significantly promoted the development of deep learning towards real-world mobile applications.

Binary neural networks (BNNs)~\cite{bnn,xnor,abcnet,bireal,bbg} push neural network quantization to the extreme. 1-bit weights and activations in BNNs can dramatically save computational cost for better real-time inference. BinaryNet~\cite{bnn} first proposed binary networks with both 1-bit weights and activations. XNOR-Net~\cite{xnor} further improved BinaryNet by introducing channel-wise scale factor of weights and activation. Dorefa-Net~\cite{dorefa} utilized layer-wise scale factor to achieve XNOR-Net like performance. ABC-Net~\cite{abcnet} enhanced the performance by using more weight bases and activation bases, but computation complexity is increased at the same time. These approaches have constantly boosted the performance of BNNs. For example, binary AlexNet in XNOR-Net~\cite{xnor} achieves a 44.2\% top-1 accuracy on the ImageNet classification task, while reducing the convolution parameters for nearly 32$\times$ over the full-precision model (56.6\% top-1 accuracy).

In constructing the binary neural network, most of existing approaches employ hard thresholding (\eg sign function) to quantize the weights and process each element independently. ``Straight through estimator'' (STE)~\cite{ste} is applied to calculate the gradient of sign function. However, the performance of the existing BNNs is far worse than that of the full-precision counterparts, \eg up to 11.6\% accuracy drop of binary AlexNet in XNOR-Net. Simply binarizing each individual weight independently does not fully explore the relationship between neurons and may not bring in the optimal solution. Moreover, estimated gradients by STE often lead to inaccurate weights that contain the noise, \ie some of the binary weights have been incorrectly flipped into the opposite values.

In this paper, we learn to binarize neurons with noisy supervision, as shown in Fig.~\ref{fig:framework}. In contrast to classical hand crafted rules to binarize the weights, we suggest a mapping from full-precision neurons to the binary ones. This mapping function can be approximated with a neural network that treats full-precision weights in a filter as a whole for the input. To help the learning of the mapping function, we take the pretrained binary weights as noisy supervisions that are close to the ideal binary neurons. An unbiased estimator is introduced for learning with noisy supervisions and avoiding noise disturbance. In an end-to-end fine-tuning, the proposed method can be a nice alternative of the sign function in BNNs by mining the relationship between neurons and taking advantage of noisy supervisions. Theoretical analysis suggests that the introduced unbiased estimator can converge to the optimal solution of binary weights under the clean distribution. Experiments on benchmark datasets including CIFAR-10 and ImageNet demonstrate that BNNs established using the proposed Learning with Noisy Supervision (LNS) method achieve state-of-the-art performance.

\section{Related Work}
In this section, we give a brief review of the related work in the field of binary neural networks and learning with noisy labels.

\subsection{Binary Neural Networks}
Binary neural networks with extremely low memory and computation cost appeal great interest from the community. Binaryconnect~\cite{binaryconnect} was proposed to train deep neural networks with binary weights. BinaryNet~\cite{bnn} further quantize both the weights and the activations to 1-bit values, starting the research on pure binary neural networks. XNOR-Net~\cite{xnor} introduce channel-wise scale factor to improve performance. Dorefa-Net~\cite{dorefa} simplifies to layer-wise scale factor and achieve similar performance with XNOR-Net. ABC-Net~\cite{abcnet} proposes to enhance the performance by using more weight bases and activation bases, but it admittedly needs more memory and computation cost than BinaryNet. There are also several works designing blocks or architectures for binary networks~\cite{bireal,shen2019searching}. Bireal-Net~\cite{bireal} introduces layer-wise identity short-cut, and AutoBNN~\cite{shen2019searching} widen or squeeze the channels in an automatic manner. All these binary models utilize STE~\cite{ste} for gradient back-propagation which would introduce inaccurate gradients for model optimization.

Some works propose new gradient calculation approach instead of STE. Bireal-Net~\cite{bireal} and DSQ~\cite{dsq} use specially designed activation function for back-propagation. PCNN~\cite{pcnn} proposes a new discrete back-propagation via projection algorithm to build BNNs. However, most of existing methods quantize each weight independently and ignore their internal relationship.

\subsection{Learning with Noisy Labels}
To learn more accurate predictions and correct biased information from noisy labels, a number of methods are proposed for learning with noisy labels, which can be divided into three categories: 

Label correction aims to correct the wrong labels in the raw labels. The existing methods usually utilize a clean label inference module to correct the noisy labels to the true ones. The inference module can be modeled by neural networks~\cite{cleannet}, graphical models~\cite{xiao2015learning}, or conditional random fields~\cite{vahdat2017toward}. However, the extra clean data or expensive noise detection process is required in these methods, which is unpractical in real-world applications.

Refined training strategies introduce new learning framework for robustness to noisy labels~\cite{mentornet,coteaching,coteaching2,wang2018iterative,tanaka2018joint}. 
These methods such as MentorNet~\cite{mentornet} and Co-teaching~\cite{coteaching,coteaching2}, change the standard learning process with complex interventions which usually need much effort to adapt and tune. 

Loss correction methods improve the standard loss function to suit for noisy labels. One common approach is modeling the noise transition matrix which defines the probability of one class flipped to another one~\cite{natarajan2013learning}. Backward and Forward~\cite{patrini2017making} introduce two alternative procedures for loss correction, provided knowing the noise transition matrix. A linear layer is added on top of the neural networks for noisy prediction correction in~\cite{goldberger2016training}. Masking~\cite{masking} derive a structure-aware probabilistic model to incorporate the structure prior. Noise robust loss function is another technique dealing with noisy labels, such as generalized cross entropy~\cite{zhang2018generalized}, label smoothing regularization~\cite{labelsmooth2}, and symmetric cross entropy~\cite{wang2019symmetric}.

Binary weights can also be recognized as prediction of a binary classifier, and the biased weights derived from the sign function are exactly the noisy label. Therefore, we present to develop a mapping that can correct the noisy binary weights and obtain BNNs with better performance.

\section{Approach}
In this section we detail the formulation of our method, including binary weight mapping model and an unbiased estimator with noisy supervision.

\subsection{Binary Weight Mapping}
In existing binary neural networks such as BinaryNet~\cite{bnn}, Bireal-Net~\cite{bireal} and Dorefa-Net~\cite{dorefa}, the weights are usually quantized with the sign function and a scale factor. In particular, the weights before quantization in a convolutional filter are denoted as $W\in\mathbb{R}^{c\times k\times k}$, where $c$ is the number of input channels, $k\times k$ is the kernel size. For simplicity in the following, we omit the scale factor, and quantized binary weights $\tilde{Q}\in\{+1,-1\}^{c\times k\times k}$ can be obtained by
\begin{equation}\label{eq:sign}
\tilde{Q}=\text{sign}(W),
\end{equation}
where $\text{sign}(\cdot)$ outputs $+1$ for positive input and $-1$ for negative input. The feature map $X\in\mathbb{R}^{n\times c\times h\times w}$ before convolution are also quantized in a similar way as the weights: $B=\text{sign}(X)$, where $B\in\{+1,-1\}^{n\times c\times h\times w}$, $n$ is the number of samples, and $h$ and $w$ are the height and weight of feature map, respectively. With the binarized weights and feature maps, the convolution computation only involves binary operations, \ie AND and POPCOUNT:
\begin{equation}\label{eq:bin-conv}
Y=B\circledast\tilde{Q},
\end{equation}
where $\circledast$ represents convolution operation with binary operations. During training, the back-propagation process of the quantization follows the straight through estimator~\cite{ste}:
\begin{equation}\label{ste}
\frac{\partial \ell_{cls}}{\partial W} \approx \text{clip}\left(\frac{\partial \ell_{cls}}{\partial \tilde{Q}},-1,1\right),
\end{equation}
where $\ell_{cls}$ is the cross entropy loss function if the neural network is for image classification, and $W$ are the latent full-precision weights to be optimized in iterations. After training, the quantized weights $\tilde{Q}=\text{sign}(W)$ will be kept for the inference.

\begin{figure}[!tb]
	\centering
	\includegraphics[width=0.8\linewidth]{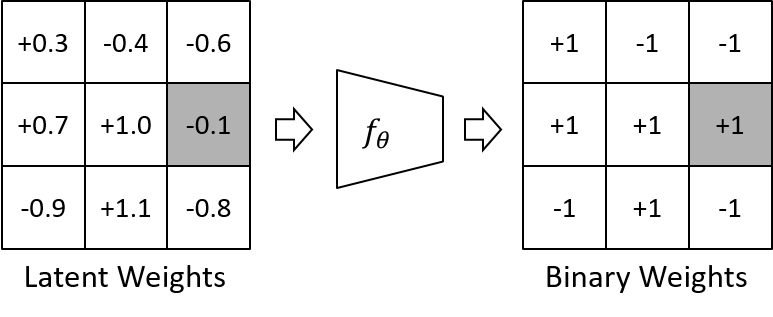}
	\vspace{-0.5em}
	\caption{Binary weight mapping. Different from simple sign function, the binarized weight here may be different from the sign of the latent weight in our method (the element in gray is an example).}
	\label{fig:transform}
	\vspace{-0.5em}
\end{figure}

The simple sign function to binarize the weights cannot take the relationship between elements into consideration and may not be the optimal. In fact, the ideal process to transform the full-precision weights to binary could be complicated and unknown. Instead of trying to fit the hand-crafted binarization rules, we propose to binarize neurons through a learned mapping function as shown in Fig.~\ref{fig:transform}. The full-precision weights are taken as a whole, and thus their internal relation can be fully explored and exploited by the mapping model to accomplish the binarization. Formally, the binarization process can be written as
\begin{equation}\label{eq:transform}
\hat{Q} = f_{\theta}(W),
\end{equation}
where $f_{\theta}$ is the mapping model with training parameters $\theta$. Compared to sign function, the binarization function in Eq.~\ref{eq:transform} is learnable and more flexible, which can approximate the binarization for the need to quantize the weights. Sign function only operates on each individual element independently, while the mapping function approximated by a neural network (Eq.~\ref{eq:transform}) can quantize each element by considering its connections with other elements.

We can embed the binary neuron mapping model in every convolutional layer which is needed to be quantized and train the entire neural network end-to-end. However, there is only the final loss (\eg $\ell_{cls}$) for supervising the entire network. The mapping model in each layer may be hard to optimize for lack of direct supervision.

\subsection{Learning with Noisy Supervision}

If the ground-truth binary weights $Q$ are provided, we can force the mapping model to learn the target under the supervision:
\begin{equation}\label{eq:gt-loss}
\ell(\hat{Q},Q)=\|\hat{Q}-Q\|_F^2,
\end{equation}
where $\|\cdot\|_F$ is the Frobenius norm of a tensor. Denoting $\hat{q}$ as each element in the predictions $\hat{Q}$, and $q$ as the corresponding ground-truth label in $Q$, Eq.~\ref{eq:gt-loss} can be represented as the following for simplicity:
\begin{equation}
\ell(\hat{Q},Q)=\|\hat{Q}-Q\|_F^2=\sum(\hat{q}-q)^2.
\end{equation}
The loss in Eq.~\ref{eq:gt-loss} is simple and easy to optimize. However, the ground-truth $Q$ is hard to obtain in practice.

If we pretrain a binary neural network as normal~\cite{dorefa}, we can easily obtain the pretrained binary model with latent weights $W$ and the corresponding binarized weights $\tilde{Q}$ in each layer. Since the gradients to $W$ in Eq.~\ref{ste} are estimated and inaccurate, the weights $W$ and the quantized $\tilde{Q}$ after optimization are also inaccurate and contain noise, that is, some of the weights have been incorrectly flipped into the opposite values. Nevertheless, as the noise corrupted $Q$, $\tilde{Q}$ still has valuable guidance to provide auxiliary supervision for learning the mapping model. Eq.~\ref{eq:gt-loss} can therefore be reformulated as
\begin{equation}\label{eq:bad-loss}
\ell(\hat{Q},\tilde{Q})=\|\hat{Q}-\tilde{Q}\|_F^2={\sum}(\hat{q}-\tilde q)^2,
\end{equation}
where $\tilde q$ is the noisy label for the mapping model. Compared to Eq.~\ref{eq:gt-loss}, the target in Eq.~\ref{eq:bad-loss} is changed to the noisy binary weights. The mapping function $f_{\theta}$ learned by minimzing the loss in Eq.~\ref{eq:bad-loss} could be seriously influenced by the noise supervision and may be harmful to the binary neural networks.

To make the full use of the noisy supervision while mitigating the influence of the noise, we seek the solution to learn from the noisy label. Inspired by the developments in noisy label learning \cite{natarajan2013learning}, we introduce a loss correction approach to avoid the noise disturbance in the mapping function learning. We view the pretrained weights $\tilde{Q}$ in BNNs as the noisy labels which is the corrupted version of $Q$. We assume that the noisy labels follow the class-conditional random noise model, 
\begin{align}
P(\tilde q=-1|q=+1) &= \rho_{+1},\\
P(\tilde q=+1|q=-1) &= \rho_{-1},
\end{align}
where $\rho_{+1}$ is the probability that the negative weight is flipped into $+1$, $\rho_{-1}$ is the probability that the positive weight is flipped into $-1$, and $\rho_{+1}+\rho_{-1}<1$. The noise rates $\rho_{+1}$ and $\rho_{-1}$ are two hyper-parameters. In binary neural networks, the number of positive weights and negative weights are similar, so the noise rate $\rho_{+1}$ and $\rho_{-1}$ should be similar as well, \ie $\rho=\rho_{+1}=\rho_{-1}$.

We aim to amend the loss function in Eq.~\ref{eq:bad-loss} to $\tilde{\ell}(\hat{Q},\tilde{Q})$ so that we have
\begin{equation}\label{eq:converge}
\mathbb{E}\left[\tilde{\ell}(\hat{Q},\tilde{Q})\right]=\ell(\hat{Q},Q),
\end{equation}
that is $\mathbb{E}\left[\tilde{\ell}(\hat{q},\tilde{q})\right]=\ell(\hat{q},q)$. Considering the cases $q=+1$ and $q=-1$ separately, we have the following equations
\begin{equation}
(1-\rho_{+1})\tilde{\ell}(\hat{q},+1) + \rho_{+1}\tilde{\ell}(\hat{q},-1) = \ell(\hat{q},+1),
\end{equation}
and 
\begin{equation}
(1-\rho_{-1})\tilde{\ell}(\hat{q},-1) + \rho_{-1}\tilde{\ell}(\hat{q},+1) = \ell(\hat{q},-1).
\end{equation}
Solving these two equations for $\tilde{\ell}(\hat{q},+1)$ and $\tilde{\ell}(\hat{q},-1)$ gives
\begin{equation}\label{eq:q+1}
\tilde{\ell}(\hat{q},+1) = \frac{(1-\rho_{-1}){l}(\hat{q},+1)-\rho_{+1}{l}(\hat{q},-1)}{1-\rho_{+1}-\rho_{-1}},
\end{equation}
and 
\begin{equation}\label{eq:q-1}
\tilde{\ell}(\hat{q},-1) = \frac{(1-\rho_{+1}){l}(\hat{q},-1)-\rho_{-1}{l}(\hat{q},+1)}{1-\rho_{+1}-\rho_{-1}}.
\end{equation}
By introducing $\tilde{q}\in\{+1, -1\}$, Eqs.~\ref{eq:q+1} and \ref{eq:q-1} can be merged into a unified loss function 
\begin{equation}\label{eq:loss}
\tilde{\ell}(\hat{q},\tilde{q}) = \frac{(1-\rho_{-\tilde{q}})\ell(\hat{q},\tilde{q})-\rho_{\tilde{q}}\ell(\hat{q},-\tilde{q})}{1-\rho_{+1}-\rho_{-1}}.
\end{equation}

Viewing each element $\tilde{q}$ as one noisy sample, we can learn a mapping model in the presence of noisy label by minimizing the sample average
\begin{equation}
\hat{f}= \arg\min_{f\in\mathcal{F}}\hat{R}_{\tilde{\ell}}(f)=\frac{1}{|\hat Q|}{\sum}\tilde{\ell}(\hat{q},\tilde{q}),
\end{equation}
where $\mathcal{F}$ can be general function class, \eg the neural networks, and $\hat{R}_{\tilde{\ell}}$ stands for the empirical $\tilde{\ell}$-risk on the observed samples. For any fixed $f\in\mathcal{F}$, the above sample average risk can converge to the $\ell$-risk under the clean distribution $D$: ${R}_{\ell,D}(f)$ even that the predictor is learned with noisy labels whereas the $\ell$-risk is computed using true labels as stated in Eq.~\ref{eq:converge}. Theoretically, the performance bound for $\hat{f}$ with respect to the clean distribution $D$ is shown in the following Theorem~\ref{theorem1}.

\begin{theorem}\label{theorem1}\cite{natarajan2013learning}
	With probability at least $1-\delta$,
	\begin{equation*}
	R_{\ell,D}(\hat{f})\leq \min_{f\in\mathcal{F}}R_{\ell,D}(f)+4L_{\rho}\mathfrak{R}(\mathcal{F})+2\sqrt{\frac{\log(1/\delta)}{2n}}.
	\end{equation*}
	where $\mathfrak{R}(\mathcal{F}):=\mathbb{E}_{X_i,\epsilon_i}[\sup_{f\in\mathcal{F}}\frac{1}{n}\epsilon_if(X_i)]$ is the Rademacher complexity of the function class $\mathcal{F}$ and $L_{\rho}\leq 2L/(1-\rho_{+1}-\rho_{-1})$ is the Lipschitz constant of the loss function $\tilde{\ell}$. Note that $\epsilon_i$'s are iid Rademacher random variables.
\end{theorem}

\begin{algorithm}[t]
	\caption{Feed-Forward and Back-Propagation Process of Binary Neuron Mapping with Noisy Supervision.} 
	\label{alg1}
	\begin{algorithmic}[1]
		\REQUIRE Pretrained latent weights $W$, input feature map $B$, the weights of mapping model $\theta$, the loss function $\mathcal{L}$, the learning rate $\eta$. 
		\STATE \textbf{Feed Foward:}
		\IF{in training stage}
		\STATE Obtain the noisy labels by sign function: $\tilde{Q} = \text{Sign}(W)$ (Eq.\ref{eq:sign});
		\STATE Compute the predicted binary weights via the mapping model: $\hat{Q}=f(W)$ (Eq.~\ref{eq:transform});
		\STATE Compute the auxiliary loss with noisy supervision: $\tilde\ell(\hat{Q},Q)=\sum\tilde{\ell}(\hat{q},\tilde{q})$ (Eq.~\ref{eq:noisy-loss}).
		\ENDIF
		\STATE Perform binary convolution: $Y=B\circledast\hat{Q}$.
		\STATE \textbf{Backward Propagation:}
		\STATE Compute the gradient of $B$: $\frac{\partial \mathcal{L}}{\partial B}=\frac{\partial \ell_{cls}}{\partial B}$, where $\frac{\partial \ell_{cls}}{\partial B}$ can be calculated as normal neural networks;
		\STATE Compute the gradient of $W$: $\frac{\partial \mathcal{L}}{\partial W}=\frac{\partial \ell_{cls}}{\partial W}+\frac{\partial \ell_i}{\partial W}$, where $\frac{\partial \ell_i}{\partial W}$ is given by Eq.~\ref{eq:grad-W} and $\frac{\partial \ell_{cls}}{\partial W}$ can be calculated as normal neural networks;
		\STATE Compute the gradient of $\theta$: $\frac{\partial \mathcal{L}}{\partial \theta}=\frac{\partial \ell_{cls}}{\partial \theta}+\frac{\partial \ell_i}{\partial \theta}$ where $\frac{\partial \ell_i}{\partial\theta}$ is given by Eq.~\ref{eq:grad-theta} and $\frac{\partial \ell_{cls}}{\partial\theta}$ can be calculated as normal neural networks;
		\STATE \textbf{Parameter Update:}
		\STATE Update the latent weights $W \leftarrow W - \eta\frac{\partial \ell}{\partial W}$, and the weights of mapping model $\theta \leftarrow \theta - \eta\frac{\partial \ell}{\partial \theta}$;
	\end{algorithmic}
\end{algorithm}

The unbiased auxiliary loss Eq. \ref{eq:loss} is therefore helpful to learn a BNN with binary neural mapping. For the $i$-th quantized layer, the auxiliary loss $\tilde{\ell}_i$ is investigated:
\begin{equation}\label{eq:noisy-loss}
\tilde{\ell}_i = \tilde\ell(\hat{Q},\tilde Q)={\sum}\tilde{\ell}(\hat{q},\tilde{q}).
\end{equation}
The introduced auxiliary loss over the predicted binary neurons from the latent full-precision weights is differential. Both the latent weights $W$ and the mapping parameters $\theta$ need to be optimized. The gradient of $\tilde{\ell}_i$ with respect to $\hat{q}$ is easy to obtain:
\begin{equation}
\begin{aligned}
\frac{\partial\tilde{\ell}_i}{\partial\hat{q}} &= \frac{(1-\rho_{-\tilde{q}})\frac{\partial\ell(\hat{q},\tilde{q})}{\partial\hat{q}} - \rho_{\tilde{q}}\frac{\partial\ell(\hat{q},-\tilde{q})}{\partial\hat{q}}}{1-\rho_{+1}-\rho_{-1}}\\
&=\frac{2(1-\rho_{-\tilde{q}})(\hat{q}-\tilde{q}) - 2\rho_{\tilde{q}}(\hat{q}+\tilde{q})}{1-\rho_{+1}-\rho_{-1}}\\
&=2(\hat{q}-\tilde{q})-\frac{4\rho_{\tilde{q}}\tilde{q}}{1-\rho_{+1}-\rho_{-1}}.
\end{aligned}
\end{equation}
Then the gradient to $\theta$ is given by
\begin{equation}\label{eq:grad-theta}
\frac{\partial\tilde{\ell}_i}{\partial\theta} = \sum_{\hat{q}}\frac{\partial\tilde{\ell}_i}{\partial\hat{q}}\frac{\partial\hat{q}}{\partial\theta},
\end{equation}
where $\frac{\partial \hat{q}}{\partial\theta}$ can be calculated with standard chain rule as $f_{\theta}(\cdot)$ is a neural network. The latent weights $W$ are the input to the mapping model $f_{\theta}(\cdot)$ to output the binary weight predictions $\hat{q}$, so the gradients are
\begin{equation}\label{eq:grad-W}
\frac{\partial\tilde{\ell}_i}{\partial W} = \sum_{\hat{q}}\frac{\partial\tilde{\ell}_i}{\partial\hat{q}}\frac{\partial\hat{q}}{\partial W}.
\end{equation}
where $\frac{\partial \hat{q}}{\partial W}$ can be calculated with back-propagation in the neural network. With the gradients of $W$ and $\theta$, the mapping model with noisy neuron correction loss $\tilde{\ell}_i$ is differential and can be optimized in an end-to-end manner.

For a binary neural network with original classification loss $\ell_{cls}$, the overall object function of our method is
\begin{equation}
\mathcal{L} = \ell_{cls}+\alpha\sum_{i}\tilde{\ell}_i,
\end{equation}
where $\alpha$ is the trade-off hyper-parameter. The proposed LNS method can be embedded into the training process of binary neural networks and trained in the end-to-end manner. The forward and back-propagation process of our method are listed in Algorithm~\ref{alg1}. After training, we obtain the optimized latent weights $W$ and mapping parameters $\theta$. We transform the latent full-precision weights to binary weights $\hat{Q}$ using the mapping neural network, and only keep those binary weights for the inference.

\section{Experiments}
In this section, we evaluate the proposed method on two image classification datasets: CIFAR-10~\cite{cifar} and ImageNet (ILSVRC12)~\cite{imagenet}, and compare our method with other BNNs.

\subsection{Datasets and Experimental Setting}
\paragraph{CIFAR-10} CIFAR-10 dataset~\cite{cifar} consists of 60,000 32$\times$32 color images belonging to 10 categories, with 6,000 images per category. There are 50,000 training images and 10,000 test images. For hyper-parameter tuning, 10,000 training images are randomly sampled for validation and the rest images are for training. Data augmentation strategy includes random crop and random flipping as in~\cite{resnet} during training. For testing, we evaluate the single view of the original image for fair comparison.

\paragraph{ImageNet} ImageNet ILSVRC 2012~\cite{imagenet} is a large-scale image classification dataset which contains over 1.2 million high-resolution natural images for training and 50k validation images in 1,000 classes. The commonly used data augmentation strategy including random crop and flipping in PyTorch examples~\cite{pytorch} is adopted for training. We report the single-crop evaluation result using $224\times224$ center crop from images.

\paragraph{Implementation Details} All the models are implemented using PyTorch~\cite{pytorch} and conducted on NVIDIA Tesla V100 GPUs. For CIFAR-10, ResNet-20 is used as baseline model. The binary baseline models are trained for 400 epochs with a batch size of 128 and an initial learning rate $0.1$. We use the SGD optimizer with the momentum of 0.9 and set the weight decay to 0. Our method is fine-tuned based on the pretrained baseline for 120 epochs using SGD optimizer. The learning rate starts from 0.01 and decayed by 0.1 every 30 epochs. For ImageNet, AlexNet and ResNet-18 are adopted for evaluation. We train the binary baseline models for 120 epochs with a batch size of 256. SGD optimizer is applied with the momentum of 0.9 and the weight decay of 0. The learning rate is set as 0.1 initially and is multiplied by 0.1 at the 70$th$, 90$th$ and 110$th$ epoch, respectively. Our method is fine-tuned from the pretrained baseline for 45 epochs with the initial learning rate 0.01 which is decayed by 0.1 every 15 epochs.

In each layer, there is a neural network for binary weight mapping. We simply use a three-layer CNN with weight shape of $2c\times c\times 3\times 3$,  $2c\times 2c\times 3\times 3$ and  $c\times 2c\times 3\times 3$, respectively. We set padding as 1 and stride as 1 in every layer of the mapping model to keep the size of output unchanged. Batch normalization and ReLU activation are inserted after the intermediate convolutional layers. The mapping model is updated for several epochs for warm start meanwhile the other weights are fixed before fine-tuning.

\subsection{Experiments on CIFAR-10}
We first conduct detailed studies on CIFAR-10 dataset for the proposed method. The widely used ResNet-20 architecture is adopted as the basic architecture, and Dorefa-Net is used as the baseline quantization method. Following the common setting in~\cite{dorefa}, all the layers except for the first convolutional layer and the last fully-connected layer for classification are quantized into 1-bit. We train the baseline binary model for 400 epochs and obtain an accuracy of 85.06\%. Based on this pretrained model, we further fine-tune with or without our method.

\begin{figure*}[!htb]
	\centering
	\small
	\renewcommand{\arraystretch}{1.0} 
	\begin{tabular}{p{0.66\columnwidth} p{0.66\columnwidth} p{0.66\columnwidth}}
		\includegraphics[width=1.0\linewidth]{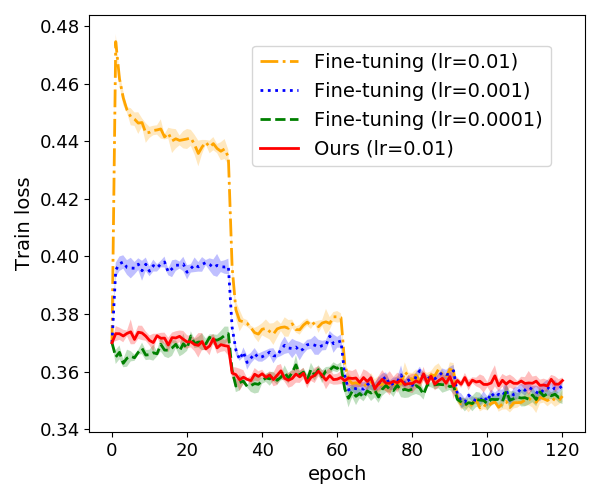} &
		\includegraphics[width=1.0\linewidth]{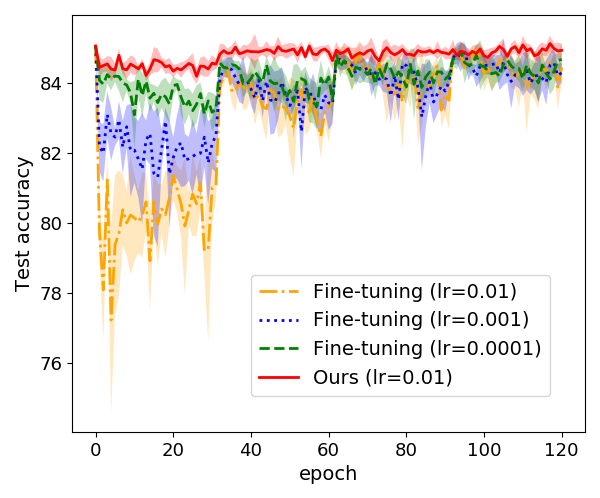} &
		\includegraphics[width=1.0\linewidth]{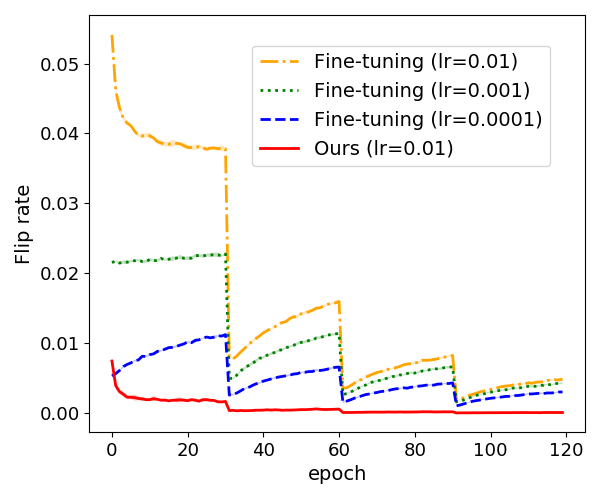}\\
		(a) Cross entropy loss \wrt epoch on CIFAR-10 train set. The mean and std values are plotted. &
		(b) Accuracy \wrt epoch on CIFAR-10 test set. The mean and std values are plotted. &
		(c) Binary weight flip rate \wrt epoch during training. The mean and std values are plotted (where std values are very small).
	\end{tabular}
	\vspace{-1em}
	\caption{Comparison of our method and simple fine-tuning (Ours with lr=0.001/0.0001 are not plotted for tidy figure as they are similar).}
	\label{fig:loss-acc}
	\vspace{-1.4em}
\end{figure*}

%

\begin{table}[htb]
	\vspace{-1em}
	\centering
	\renewcommand{\arraystretch}{1.05} 
	\caption{Ablation study on CIFAR-10. For the compared methods, we run them 5 times and show ``best (mean$\pm$std)''.}\label{tab:ablation}
	\vspace{0.2em}
	\small
	\setlength{\tabcolsep}{3pt}{
		\begin{tabular}{c||c}
			\hline
			Method & Acc (\%) \\
			\hline\hline
			Dorefa-Net~\cite{dorefa} (Baseline) & 85.06 \\
			Fine-tuning & 85.32 (85.26$\pm$0.06) \\
			Ours w/o Noisy-supervision & 85.43 (85.36$\pm$0.06) \\
			LNS (Ours) & 85.78 (85.56$\pm$0.11) \\
			\hline
		\end{tabular}
	}
	\vspace{-1.2em}
\end{table}

\paragraph{Effectiveness of Our Method.} To verify the effectiveness of our method, we fine-tune the baseline model without mapping model and our method for 120 epochs with all the same experimental settings as stated in implementation details. In our method, we setting the hyper-parameters as $\alpha=1.0$ and $\rho=0.005$. We run them 5 times and show the best, mean and standard values in Table~\ref{tab:ablation}. After fine-tuning using our method, we can see that our method without noisy supervision achieves a mean accuracy of 85.36\%, adding noisy supervision further improve the accuracy to 85.56\%, while simply fine-tuning achieves 85.26\%. Both simple fine-tuning and our method can improve the baseline model, but the performance of our method is much better than simple fine-tuning. The results indicate the effectiveness of the proposed binary neuron mapping and the corresponding noisy supervision. The highest accuracy of our method can achieve 85.78\%, which is the state-of-the-art as shown in the latter analysis.

We also plot the loss curve and accuracy curve to observe the effect of our method during training. The cross entropy loss curves of simple fine-tuning and our method are shown in Fig.~\ref{fig:loss-acc}(a), and the test accuracy curves of them are shown in Fig.~\ref{fig:loss-acc}(b). The initial loss value and accuracy are 0.37 and 85.06\%, respectively, from the pretrained baseline model. At first, we find that Fine-tuning has a much larger loss than our method with the same initial learning rate (lr=0.01), so we decrease the learning rate. Although the loss in Fine-tuning (lr=0.001/0.0001) is decreased, the accuracy on test set has no improvement. From Fig~\ref{fig:loss-acc}(a), we can see that the simple fine-tuning changes the loss at the first as it disturbs the pretrained binary weights largely, while our method does not change the loss much as the mapping model only changes a small portion of the weights. When we decrease the learning rate at the 30$th$, 60$th$ and 90$th$ epoch, the loss values in Fine-tuning and our method will have a relatively large drop. At the last several epochs, simple fine-tuning method with different learning rate achieve a smaller train loss, but a lower test accuracy than our method (Fig.~\ref{fig:loss-acc}(b)). This means that our method can alleviate over-fitting by imposing a noisy supervision on each layer as the noisy weights are often those that over-fit the training data.


From the loss curve and accuracy curve, we know that the training process of our method is more stable than that of simple fine-tuning method. We show the flip rate of the binary weights after each epoch in Fig.~\ref{fig:loss-acc}(c), where flip rate means the ratio of binary weights that are flipped into the opposite values. The flip rate decrease gradually during training in all these curves. The flip rate in Fine-tuning is always higher than our method, which means our method only change a small portion of binary weights which are likely to be noise to explore better performance.

\begin{figure}[!htb]
	\vspace{-0.5em}
	\centering
	\includegraphics[width=0.95\linewidth]{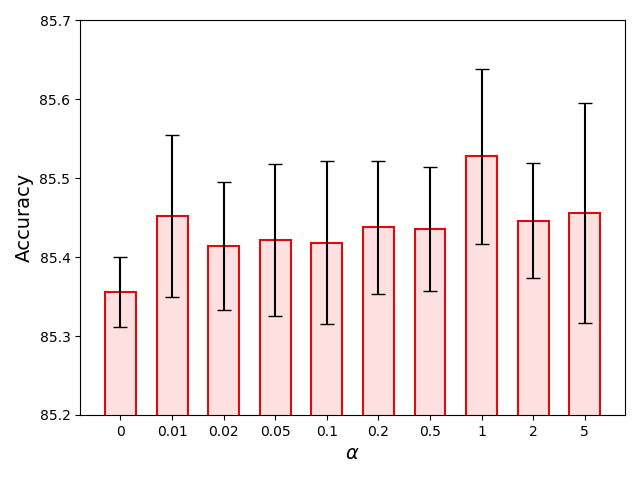}
	\vspace{-1.0em}
	\caption{Accuracy v.s. $\alpha$ on CIFAR-10 val set.}
	\label{fig:alpha}
	\vspace{-1em}
\end{figure}

\begin{figure}[!htb]
	\centering
	\includegraphics[width=0.95\linewidth]{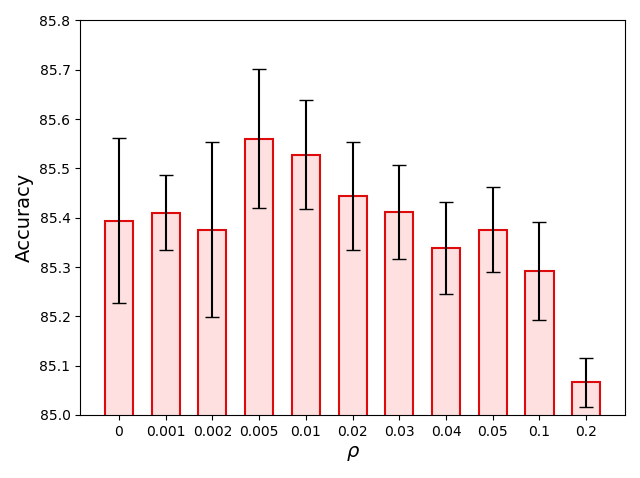}
	\vspace{-1.0em}
	\caption{Accuracy v.s. $\rho$ on CIFAR-10 val set.}
	\label{fig:rho}
	\vspace{-1em}
\end{figure}

\paragraph{Analysis of Hyper-parameters.}
There are two hyper-parameters in our method, \ie $\alpha$ for balancing the cross entropy classification loss and the noisy neuron correction loss, and $\rho$ for controlling the noise rate in the binary weight transformation. We run all the models 5 times and report the mean and std values of the accuracy on CIFAR-10 validation set.

We first fix $\rho=0.01$ and tune $\alpha$ in range of $\{0,0.01,$ $0.02,0.05,0.1,0.2,0.5,1,2,5\}$ to see the influence of $\alpha$. The results are shown in Fig.~\ref{fig:alpha}. The mean accuracy is 85.35\% when $\alpha=0$, which is higher than the simple fine-tuning. This verifies the effectiveness of the mapping model without self-supervision. When we increase the value of $\alpha$, the accuracy is improved over that at $\alpha=0$. The highest mean accuracy occurs around $\alpha=1$, \ie 85.53\%. We can see that our method works at a large range of $\alpha$ and can choose $\alpha$ around 1 for the best performance on CIFAR-10.

For the noise rate $\rho$, we fix $\alpha$ as 1 and test $\rho$ in $\{0,$ $0.001,0.002,0.005,0.01,0.02,0.03,0.04,0.05,0.1,0.2\}$. From the results in Fig.~\ref{fig:rho}, we can see that the mean accuracy at different noise rate is different. When $\rho$ is small (around 0.001), the mean accuracy is about 85.4\%, which is higher than baseline and simple fine-tuning method. Our method achieve the best mean accuracy at $\rho=0.005$, which means the ground-truth noise rate is about 0.5\%. When $\rho$ is too large, such as $\rho=0.2$, the mean accuracy drops significantly, even below the simple fine-tuning method. This is due to that there are not so much noise in the pretrained binary weights, setting $\rho$ too large disturbs the good weights and is harmful to the performance.

\begin{table}[htb]
	\vspace{-1em}
	\centering
	\renewcommand{\arraystretch}{1.05} 
	\caption{Comparison with SOTA on CIFAR-10. Dorefa-Net and XNOR-Net are implemented by ourselves. For our method, we run it 5 times and show ``best (mean$\pm$std)'' as in \cite{resnet}.}\label{tab:cifar10}
	\vspace{0.2em}
	\small
	\setlength{\tabcolsep}{3pt}{
		\begin{tabular}{c||c}
			\hline
			Method & Acc (\%) \\
			\hline\hline
			Dorefa-Net~\cite{dorefa} (Baseline) & 85.06 \\
			XNOR-Net~\cite{xnor} & 85.23 \\
			TBN~\cite{tbn} & 84.34 \\
			DSQ~\cite{dsq} & 84.11 \\
			LNS (Ours) & 85.78 (85.56$\pm$0.11) \\
			\hline
		\end{tabular}
	}
	\vspace{-1em}
\end{table}

\paragraph{Comparison with SOTA.}
We compare our method with some other state-of-the-art binary neural networks such as Dorefa-Net~\cite{dorefa}, XNOR-Net~\cite{xnor}, and DSQ~\cite{dsq}. The results of comparison are listed in Table~\ref{tab:cifar10}. Note that only the best accuracy is reported for other methods. From the results, our method outperforms the competitors by a large margin and achieves the state-of-the-art result (85.78\% accuracy).

\subsection{Experiments on ImageNet}
In order to validate our method on large-scale dataset, we conduct more experiments on ImageNet classification dataset. Two common network architectures, \ie AlexNet~\cite{alexnet} and ResNet-18~\cite{resnet}, are used for experiments.

\paragraph{Effectiveness of Our Method.}
We test the effect of our method by deploying the proposed method on ResNet-18. The baseline here is the binary ResNet-18 which is quantized using Dorefa-Net~\cite{dorefa}. We fine-tune the pretrained binary ResNet-18 with or without our method for 60 epochs. These two models use the same hyper-parameter settings. The simple fine-tuning without using our method can improve the Top-1 accuracy to 52.8\%. Our method achieves 53.1\%, much higher than the baseline and the simple fine-tuning.

\begin{table*}[htb]
	\centering
	\renewcommand{\arraystretch}{1.05} 
	\caption{Comparison with SOTA of ResNet-18 architecture on ImageNet. `W' and `A' refer to the weight and activation bitwidth, respectively. $^{\dag}$ represents the result from our implementation.}\label{tab:res18-b}
	\vspace{0.2em}
	\small
	\setlength{\tabcolsep}{13pt}{
		\begin{tabular}{c||c|c|c|c|c|c}
			\hline
			Method & W & A & Memory& FLOPs & Top-1 & Top-5 \\
			\hline\hline
			ResNet-18~\cite{resnet} & 32 & 32 & 374 Mbit & 1810 M & 69.6\% & 89.2\% \\
			BWN~\cite{xnor}  & 1 & 32 & 34 Mbit & 975 M & 60.8\% & 83.0\% \\
			HWGQ~\cite{hwgq}  & 1 & 2 &  34 Mbit & 193 M & 59.6\% & 82.2\% \\
			TBN~\cite{tbn}  & 1 & 2 &  34 Mbit & 193 M & 55.6\% & 79.0\% \\
			BinaryNet~\cite{bnn}  & 1 & 1 & 28 Mbit & 149 M & 42.2\% & 67.1\% \\
			Dorefa-Net~\cite{dorefa}$^{\dag}$ & 1 & 1 & 34 Mbit & 163 M & 52.5\% & 76.7\% \\
			XNOR-Net~\cite{xnor} & 1 & 1 &  34 Mbit & 167 M & 51.2\% & 73.2\% \\
			Bireal-Net~\cite{bireal}  & 1 & 1 &  34 Mbit & 163 M & 56.4\% & 79.5\% \\
			Bireal-Net~\cite{bireal}+PReLU (Baseline)  & 1 & 1 & 34 Mbit & 163 M & 59.0\% & 81.3\% \\
			PCNN ($J$=1)~\cite{pcnn}   & 1 & 1 & 34 Mbit & 167 M & 57.3\% & 80.0\% \\
			Quantization networks~\cite{quantization-net} & 1 & 1 & 34 Mbit & 163 M & 53.6\% & 75.3\% \\
			Bop~\cite{bop}  & 1 & 1 & 34 Mbit & 163 M & 54.2\% & 77.2\% \\
			GBCN~\cite{gbcn} & 1 & 1 &  34 Mbit & 167 M & 57.8\% & 80.9\% \\
			IR-Net~\cite{irnet} & 1 & 1 &  34 Mbit & 163 M & 58.1\% & 80.0\% \\
			LNS (Ours) & 1 & 1 &  34 Mbit & 163 M & 59.4\% & 81.7\% \\
			\hline
		\end{tabular}
	}
	\vspace{-1em}
\end{table*}

\begin{table}[htb]
	\vspace{-0.5em}
	\centering
	\renewcommand{\arraystretch}{1.05} 
	\caption{Effectiveness of our method on ResNet-18 architecture on ImageNet.}\label{tab:res18-a}
	\vspace{0.2em}
	\small
	\begin{tabular}{c||c|c}
		\hline
		Method & Top-1 & Top-5 \\
		\hline\hline
		Dorefa-Net~\cite{dorefa} (Baseline) & 52.5\% & 76.7\% \\
		Fine-tuning & 52.8\% & 76.8\% \\
		LNS (Ours) & 53.1\% & 77.0\% \\
		\hline
	\end{tabular}
	\vspace{-1.5em}
\end{table}

\paragraph{Comparison with SOTA.}
While the ablation study has evaluated the effectiveness of the proposed method, we also compare our method with the state-of-the-art methods to show the superiority of our method. The compared binary neural network methods include BinaryNet~\cite{bnn}, Dorefa-Net~\cite{dorefa}, XNOR-Net~\cite{xnor}, Bireal-Net~\cite{bireal}, PCNN~\cite{pcnn}, Bop~\cite{bop}, GBCN~\cite{gbcn}, \etc. Two representative 2-bit neural networks, \ie HWGQ~\cite{hwgq} and TBN~\cite{tbn}, are also included. Following the common settings~\cite{bnn,bireal}, we do not quantize the first convolutional layer and the last fully connected layer for classification.

In ResNet-18 experiments, except for BinaryNet and ABC-Net, all the other methods including our method do not quantize the down-sample layers for fair comparison. The statistics of the compared methods are listed in Table~\ref{tab:res18-b}. The FLOPs are calculated as real-valued floating-point multiplication plus 1/64 of the amount of 1-bit multiplication as the binary operations including AND and POPCOUNT can be performed in a parallel of 64 by the mainstream CPUs~\cite{bireal}. We use the ResNet-18 architecture in Bireal-Net as baseline and insert PReLU activation~\cite{prelu} after every binary convolutional layer. This strong baseline has a Top-1 accuracy of 59.0\%. Our method is fine-tuned based on the pretrained baseline and finally achieve 59.4\% Top-1 and 81.7\% Top-5 accuracies, which are higher than the compared models and achieve the state-of-the-art results for binary ResNet-18. It is encouraging to see that our method can beat some methods with 2-bit activations, such as HWGQ~\cite{hwgq} and TBN~\cite{tbn}. This gives us the confidence to achieve higher performance with lower bit-width in neural networks.

We also compare our method with several state-of-the-art models for AlexNet architecture which does not has residual connections. from the results in Table~\ref{tab:alexnet}, tt can be seen that our method outperforms the compared models such as BinaryNet~\cite{bnn}, and Dorefa-Net~\cite{dorefa}, which validates the superiority of our method for different architectures. Moreover, our method with only layer-wise scale factor can achieve higher Top-1 accuracy than XNOR-Net~\cite{xnor} which uses channel-wise scale factor and more parameters.

\begin{table}[htb]
	\vspace{-1em}
	\centering
	\renewcommand{\arraystretch}{1.05} 
	\caption{Comparison with SOTA of AlexNet architecture on ImageNet. $^{\dag}$Results from our implementation.}\label{tab:alexnet}
	\vspace{0.2em}
	\small
	\setlength{\tabcolsep}{5pt}{
		\begin{tabular}{c||c|c}
			\hline
			Method & Memory & Top-1 \\
			\hline\hline
			FP32-AlexNet~\cite{alexnet} & 1860 Mbit & 56.6\%  \\
			BinaryNet~\cite{bnn} & 180 Mbit & 41.8\% \\
			Dorefa-Net~\cite{dorefa} & 180 Mbit & 43.6\% \\
			Dorefa-Net (Baseline)$^{\dag}$ & 180 Mbit & 43.9\% \\
			XNOR-Net~\cite{xnor} & 181 Mbit & 44.2\% \\
			LNS (Ours) & 180 Mbit & {44.4\%} \\
			\hline
		\end{tabular}
	}
	\vspace{-1.0em}
\end{table}

\section{Conclusion}
In this paper, we have presented a novel binary neuron mapping method with noisy supervision that leads to state-of-the-art performance for binary neural networks. We apply a learnable mapping model instead of the sign function for weight quantization. The unbiased estimator of mean square error loss is applied to learn from the pretrained binary models. We show that the specially designed loss can converge to $\ell$-risk under the clean distribution of binary weights. The experiments on various datasets and neural architectures have verified the effectiveness of the proposed method. The resulted binary neural networks achieve the state-of-the-art performance compared with other approaches.

\clearpage

\section*{Acknowledgement}
The authors thank the anonymous reviewers for their helpful comments in revising the paper. This work was supported in part by National Key R\&D Program of China (2017YFB1002701), NSFC (61632003,61672502), Macau S\&T Development Fund (0018/2019/AKP), UM Research Fund (MYRG2019-00006-FST), and in part by the Australian Research Council under Project DE-180101438.

\bibliography{ref}
\bibliographystyle{icml2020}

\end{document}